\documentclass[sigconf]{acmart}
\AtBeginDocument{%
  }
\copyrightyear{2023}
\acmYear{2023}
\setcopyright{rightsretained}
\acmConference[KDD '23]{Proceedings of the 29th ACM SIGKDD Conference on Knowledge Discovery and Data Mining}{August 6--10, 2023}{Long Beach, CA, USA}
\acmBooktitle{Proceedings of the 29th ACM SIGKDD Conference on Knowledge Discovery and Data Mining (KDD '23), August 6--10, 2023, Long Beach, CA, USA}
\acmDOI{10.1145/3580305.3599907}
\acmISBN{979-8-4007-0103-0/23/08}

\allowdisplaybreaks
\usepackage{tablefootnote}
\usepackage{array}

\usepackage[inline]{enumitem}
\usepackage[capitalize,noabbrev,nameinlink]{cleveref}
\usepackage{pmboxdraw} 
\usepackage{caption}
\usepackage{balance}
\usepackage{anyfontsize}
\usepackage{scalerel,graphicx,xparse}
\usepackage{wasysym}
\NewDocumentCommand\laughcry{}{\scalerel*{\includegraphics{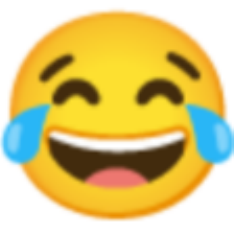}}{▁}}
\NewDocumentCommand\anxious{}{\scalerel*{\includegraphics{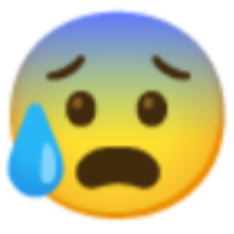}}{▁}} 

\begin{document}

\title{SMILE: Evaluation and Domain Adaptation for \\ Social Media Language Understanding}
\author{Vasilisa Bashlovkina}
\email{vasilisa@google.com}
\affiliation{%
  \institution{Google Research}
   \city{New York}
   \state{NY}
   \country{USA}
}
\author{Riley Matthews}
\email{rileymatthews@google.com}
\affiliation{%
  \institution{Google Research}
   \city{New York}
   \state{NY}
   \country{USA}
}
\author{Zhaobin Kuang}
\email{kuangz@google.com}
\affiliation{%
  \institution{Google Research}
   \city{New York}
   \state{NY}
   \country{USA}
}
\author{Simon Baumgartner}
\email{simonba@google.com}
\affiliation{%
  \institution{Google Research}
   \city{New York}
   \state{NY}
   \country{USA}
}
\author{Michael Bendersky}
\email{bemike@google.com}
\affiliation{%
  \institution{Google Research}
   \city{Mountain View}
   \state{CA}
   \country{USA}
}
\renewcommand{\shortauthors}{Vasilisa Bashlovkina, Riley Matthews, Zhaobin Kuang, Simon Baumgartner, \& Michael Bendersky}

\begin{abstract}
We study the ability of transformer-based language models (LMs) to understand social media language. Social media (SM) language is distinct from standard written language, yet existing benchmarks fall short of capturing LM performance in this socially, economically, and politically important domain. We quantify the degree to which social media language differs from conventional language and conclude that the difference is significant both in terms of token distribution and rate of linguistic shift. Next, we introduce a new benchmark for Social MedIa Language Evaluation (SMILE\smiley) that covers four SM platforms and eleven tasks. Finally, we show that learning a tokenizer and pretraining on a mix of social media and conventional language yields an LM that outperforms the best similar-sized alternative by 4.2 points on the overall SMILE\smiley~ score.
\end{abstract}


\maketitle

\section{Introduction}
\label{sec:intro}
Social media (SM) plays an increasingly important role in our lives. As of 2021, seven out of ten US adults use at least one social media platform like Facebook, Twitter, Instagram, or Pinterest \citep{auxier2021social}.
That proportion likely underestimates SM use if we broaden the definition of SM to include user-generated content like restaurant reviews, news article comments, and forum discussions. 
The ever-growing trove of text produced by social media users is both a challenge and an opportunity for natural language processing (NLP). NLP models with a strong grasp of social media language could perform a variety of socially, economically, and politically important tasks. They could, for example, tackle automatic content moderation to improve the quality of online discourse, summarize restaurant reviews to simplify the decision-making process of hungry customers, and detect and stunt disinformation campaigns aimed at sowing societal instability.

However, modeling social media language is inherently challenging. Due to its informal, noisy, and fast-evolving nature, social media language on platforms such as Twitter \citep{deluciabernice} is different from the language found in books, news publications, and Wikipedia.  
Additionally, the tasks that organically arise from the social media domain (trend detection, emoji prediction, cyberbullying detection, online marketing, etc.) are qualitatively different from the tasks natural to the domain of standard written language (translation, entailment, grammar checking, etc.). 
Although the recipe of pretraining language models on massive conventional corpora has been successful in pushing the state-of-the-art of general language understanding \citep{devlin2018bert, raffel2020exploring, brown2020language, chowdhery2022palm, bommasani2021opportunities, liu2019roberta}, it is unclear if this recipe's success will transfer to the social media domain. 
The reason for this lack of clarity is that general language understanding benchmarks \citep{wang2018glue, wang2019superglue, talukder2019bengali, rajpurkar2016squad, bojar2014findings, liang2022holistic, srivastava2022beyond}  include neither SM data nor tasks and therefore do not measure social media language understanding. 

While huge strides have been made in the social media language understanding literature to mitigate these challenges, existing work has spotty coverage of the full range of social media platforms and social media language understanding tasks. First, some related works studied vocabulary shift on a single platform (Twitter,  \citet{amba2021dynamic}) and compared corpora of different genres \citep{fothergill2016evaluating, lu2021diverging}. However, they did not directly compare social media language with conventional language. Similarly, many existing social media language understanding benchmarks either focus on a single platform  \citep{barbieri2020tweeteval, zhang2015character} or a single task \citep{demszky2020goemotions, kim2018abstractive, borkan2019nuanced} and thus cannot provide a holistic evaluation. Finally, while there exist language models pretrained on multiple platforms in specialized domains such as scientific literature \citep{beltagy2019scibert},  biomedical text \citep{gu2021domain, luo2022biogpt}, and electronic health records \citep{yang2022large}, some of the most popular social media language models \citep{nguyen2020bertweet, barbieri2020tweeteval, ji2021mentalbert, deluciabernice, zhang2022twhin, barbieri2022xlm} are explicitly pretrained on data from a single social media platform. It is unclear whether these models generalize well \textit{across} social media platforms.

Towards a more comprehensive understanding of social media language across multiple platforms and multiple application scenarios, we propose a new benchmark and a recipe for training language models that accounts for the divergence between social media language and conventional language. Specifically, we consider social media language understanding \emph{in English} and make the following contributions: 
\begin{itemize}[leftmargin=*]

\item We conduct a time-aligned comparison between the vocabulary (token) distribution of (1) posts from Twitter and Reddit and (2) that of mC4 \citep{xue2020mt5}, a conventional text corpus used to pretrain language models in many existing works \citep{xue2022byt5, xue2020mt5, tay2021charformer}. We observe a substantial difference between the two distributions and also find that social media language changes twice as fast as conventional language (\cref{sec:compare}).

\item We compile a Social MedIa Language Evaluation (SMILE\smiley) benchmark that includes social media language data from four platforms (Twitter, Reddit, Yelp, and Civil Comments) across both classification and generation tasks organically arising from the social media domain.
This newly compiled benchmark coupled with an evaluation protocol is a well-rounded toolkit for the evaluation of an LM's social media language understanding (\cref{sec:benchmark}). 

\item We provide an effective recipe for training LMs for social media language understanding backed by a large-scale empirical study conducted using the SMILE\smiley~ benchmark and training regimen for T5-based architectures \citep{raffel2020exploring}. Our study suggests that by training a custom tokenizer and pretraining the model from scratch using a corpus of both social media and conventional language, we can improve performance by 4.2 points compared to a similarly-sized baseline model (\cref{sec:model}). We carry out additional ablation studies in \cref{sec:ablation}.
\end{itemize}

It is worth noting that very large language models (LLMs) like PaLM \citep{chowdhery2022palm,thoppilan2022lamda} exhibit emergent few-shot capabilities in understanding and generating different styles, modes, and dialects of language. Thus the challenges of social media language understanding outlined above may be less pronounced for this class of models. To that end, we outline some interesting research directions in \cref{sec:future_work}. The central focus of our study, however, is models that are three orders of magnitude smaller in terms of parameters---i.e. 220M vs. 540B for the largest PaLM model---and follow a pretraining-fine-tuning workflow (\cref{sec:background}) rather than the few-shot learning setup.

The rest of the paper is organized as follows. \cref{sec:background} provides necessary background. \cref{sec:compare} presents our comparison between social media language and conventional language. In \cref{sec:benchmark}, we describe the compilation of the SMILE\smiley~ benchmark. Using the SMILE\smiley~ benchmark, we report the findings of our empirical study on learning language models for social media language understanding in \cref{sec:model}. We carry out related ablation studies in \cref{sec:ablation} and discuss future work in \cref{sec:future_work}. Finally, we conclude the paper in \cref{sec:conclusion}.

\section{Background}
\label{sec:background}

We describe a typical end-to-end workflow for training models for language understanding. This workflow proceeds in three steps: data preparation, tokenizer setup, and model training. We present these three steps in detail, describing related works and highlighting why they may be insufficient for social media language understanding.

\paragraph{Data Preparation} 
In this step, we gather data needed to produce a language model from scratch. Two types of data are needed: training corpora and evaluation benchmarks. 
\begin{itemize}[leftmargin=*]
\item \emph{Training corpora} typically consist of text crawled from the web with filters applied for data quality. For example, the C4 corpus \citep{raffel2020exploring}, which is widely used to pretrain language models for general language understanding, contains 750GB of meticulously filtered web text free from offensive language, paragraphs that are too short, and sentences that do not end in a terminal punctuation mark. While such filters may make sense for general language understanding, it is unclear whether these design choices are as suited to social media language understanding: units of social media text tend to be short and poorly punctuated, and understanding offensive language likely key to downstream social media tasks like online safety. In fact, as a result of filtering, the C4 corpus does not include any Twitter data. We further study this disparity between social media language and conventional language in \cref{sec:compare}.

\item \emph{Evaluation benchmarks} typically contain a number of small-scale datasets with labels for a particular task, e.g., sentiment classification. These datasets typically have splits: a train split used to fine-tune a model and a test split used to evaluate the fine-tuned model's performance on that task. In social media understanding, a popular evaluation benchmark is TweetEval \citep{barbieri2020tweeteval}, which consists of seven classification tasks measuring the capacity of language models to understand tweets. Other evaluation benchmarks exist  \citep{barbieri2022xlm, aluru2021deep}; however, existing benchmarks are skewed towards particular social media platforms or particular tasks. We seek to provide a more comprehensive measure with our newly-complied benchmark in \cref{sec:benchmark}.
\end{itemize}

\paragraph{Tokenizer setup} Tokenization is the process of segmenting strings of text into sequences of characters known as tokens. The set of possible tokens is determined in advance and constitutes the vocabulary of the language model, the basic units of the model's language understanding. Our study relies on the SentencePiece model (SPM) \citep{kudo2018sentencepiece}, which selects frequently observed whitespace-agnostic tokens from a large training corpus. Given the discrepancy between social media language and the standard language of the web, one may conjecture that training a tokenizer specific to social media language could be helpful. However, related results from existing works in learning domain-specific language models are mixed \citep{gu2021domain,barbieri2020tweeteval}. In \cref{sec:model} we investigate this conjecture further.

\paragraph{Model training} Pioneered by \citet{devlin2018bert}, masked language modeling (MLM) is an effective procedure of pretraining transformer based language models \citep{vaswani2017attention}. The setup asks the model to predict randomly masked tokens in sentences drawn from the pretraining corpus in a Cloze test fashion. Once the model is pretrained, it is further trained (i.e.~fine-tuned) on the evaluation benchmark train split. The fine-tuned model is then evaluated on the benchmark test split to measure its performance on a particular task. There are a few existing language models for social media language understanding. For example,  \citet{barbieri2020tweeteval, nguyen2020bertweet, deluciabernice, zhang2022twhin, barbieri2022xlm} reported language models trained on Twitter text. Meanwhile, MentalBERT \citep{ji2021mentalbert}, a langauge model for mental-health-related tasks, is trained on Reddit data. In general language understanding, \citet{thoppilan2022lamda} and  \citet{chowdhery2022palm} use 50\% social media content in pretraining LaMDA and PaLM, respectively. As mentioned in \cref{sec:background}, this class of models is not in scope for this study.

\begin{figure}
    \centering
    \includegraphics[width=1.0\linewidth]{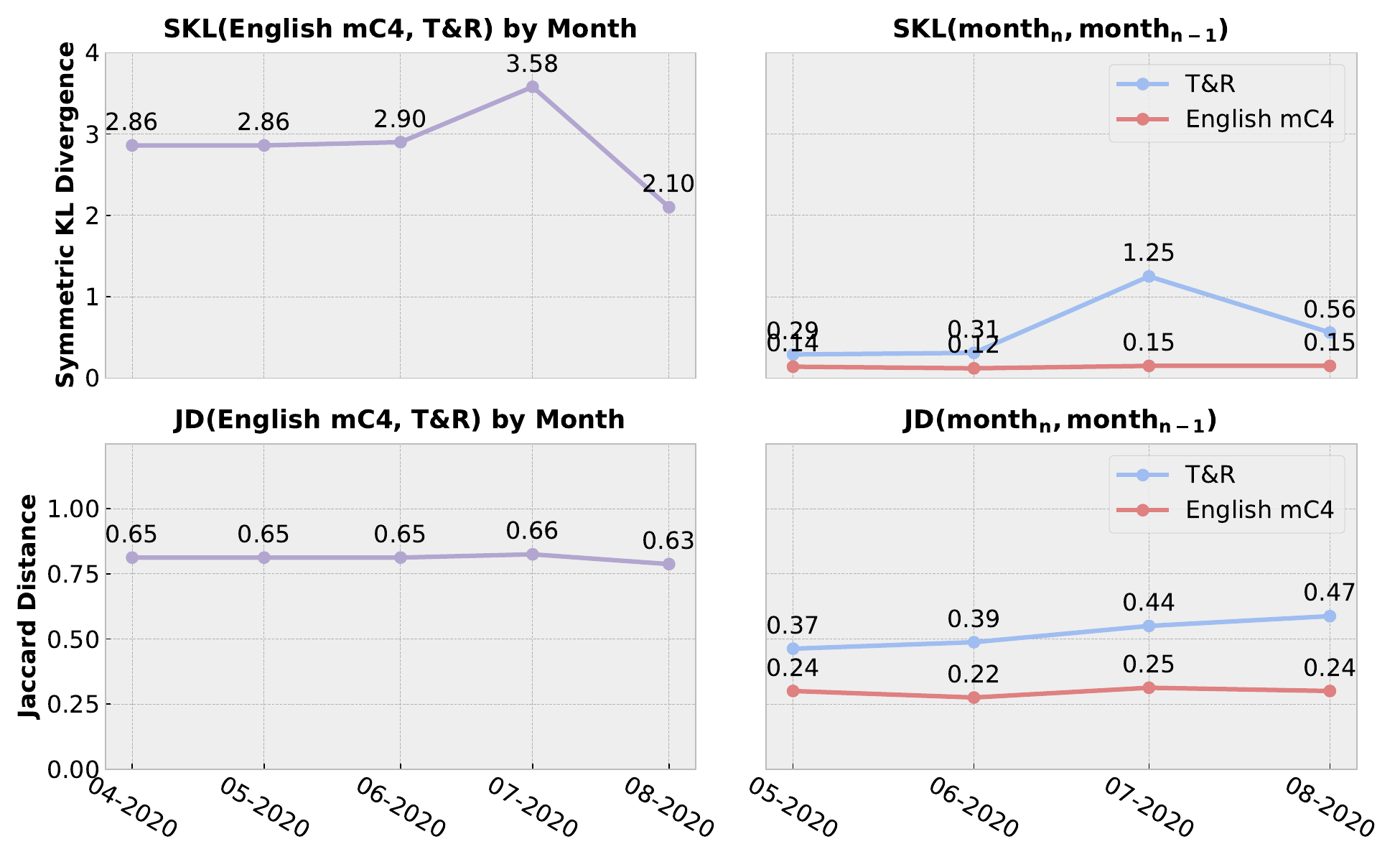}
    \caption{Left: Symmetric KL (SKL) divergence and Jaccard distance (JD) between the vocabulary distributions of  T\&R and English mC4. Right: SKL and JD of the month-to-month vocabulary distributions of T\&R and English mC4.}
    \label{fig:graphs}
    \vspace{-1.7em}
\end{figure}

\section{Comparing Social Media Language with Conventional Language}
\label{sec:compare}

In this section, we empirically study the difference between conventional and social media language to motivate our subsequent exploration of model training recipes. Our findings suggest that (1) the language used on social media is significantly different from the standard language of the web (\cref{sec:v-diff}) and (2) social media language changes twice as fast as conventional language (\cref{sec:v-shift}). 

\paragraph{Datasets} To represent social media language, we compile a collection of English-only text posts from publicly-crawlable Twitter (80\%) and Reddit (20\%) pages (T\&R). To represent conventional language we use the English portion of mC4, C4's multilingual descendant that is also widely used to pretrain language models for general language understanding \citep{xue2020mt5}. We focus on the subset of T\&R and English mC4 that overlap in post/document publish time: between April 2020 and August 2020. We then split the two corpora into five segments representing each month, sampling 4M posts/documents from each split.

\subsection{Vocabulary Difference between Conventional and Social Media Language}
\label{sec:v-diff}
We quantify the difference in social media language and conventional language by measuring the difference in token distribution between T\&R and the English mC4 corpus.

\paragraph{Metrics} We compute the symmetric KL divergence (SKL, a.k.a.~population stability index or PSI) between token distributions in each five monthly segments of T\&R and English mC4 as a measure of their difference. SKL ranges from $0$ to $+\infty$. \citet{yurdakul2020statistical} suggests that, as a rule of thumb, an SKL smaller than $0.1$ implies little difference, an SKL between $0.1$ and $0.25$ implies moderate difference, and an SKL higher than $0.25$ implies a significant difference between two distributions. We use this rough rule of thumb to interpret the SKL divergence results. Following a related study \citep{amba2021dynamic}, we also report Jaccard distance (JD) between each two vocabularies (token sets). Jaccard distance ranges from $0$ to $1$. While various other approaches are proposed in computational linguistics \citep{fothergill2016evaluating, kilgarriff1997using, dunn2022natural, kilgarriff2001comparing} to compare corpora, we focus on the comparison between token distributions because our downstream workflow takes tokens as input, and hence the changes in token distribution directly influence subsequent language modeling.

\paragraph{Protocol}
For each of the five month-segments of T\&R and English mC4, we train an SPM tokenizer with a vocabulary size of 50,000 to compute the token frequency distribution for that subcorpus. For each segment of T\&R and segment of English mC4 corresponding to the same month, we take the tokens that are in either vocabulary along with their frequencies to obtain two categorical distributions over the same set of tokens. We then calculate the SKL divergence between these two distributions. To calculate the Jaccard distance, we count the number of tokens in the intersection $i$ and union $u$ of the two segments of the same month from T\&R and English mC4. The Jaccard distance is then given as $1-i/u$. Note that our time-aligned comparison controls 
for temporal confounding, hence providing a more accurate perspective on the difference between T\&R and English mC4. 

\paragraph{Results}  \cref{fig:graphs} reports the token distribution difference between T\&R and the English mC4 corpus over five months. In terms of SKL, the divergence between the two is consistently higher than $2.8$, a significant difference. The consistently high Jaccard distance ($>0.65$) between T\&R and English mC4 corroborates this finding. Thus we conclude that social media language is significantly different from conventional language. These empirical observations validate our intuition and motivate the need for a social media language modeling recipe distinct from the status quo.
\subsection{Temporal Vocabulary Shift in Social Media Language and Conventional Language}
\label{sec:v-shift}
We quantify the temporal vocabulary shifts in social media and in conventional language by measuring the month-to-month token distribution shift in T\&R and English mC4 and compare the rate at which these two modes of language change over time.

\paragraph{Metrics and Protocol} For both T\&R and English mC4, we compute SKL divergence and Jaccard distance between the token distributions of adjacent months to measure. The calculation of these metrics follows the same protocol as in \cref{sec:v-diff}. The five monthly segments result in four month-to-month SKL divergence and Jaccard distance statistics for both T\&R and English mC4.

\paragraph{Results} \cref{fig:graphs} presents the month-to-month SKL and Jaccard distance for T\&R and English mC4. In terms of SKL, we observe significant month-to-month changes ($>0.25$) of the vocabulary in T\&R. Meanwhile, the changes in the English mC4 corpus are less significant ($<0.25$). In terms of both month-to-month SKL divergence and Jaccard distance, the numbers for T\&R are roughly twice as high as for English mC4. These observations suggest that social media language changes faster than conventional language.

\section{The Social Media Language Evaluation (SMILE) Benchmark}
\begin{table}
\centering
 \caption{Key characteristics of the social media language understanding tasks in the SMILE Benchmark; CLS: classification task; GEN: generation task.}
    \label{tab:benchmarks}
\resizebox{\columnwidth}{!}{
\begin{tabular}{||l | c | c | c c | c ||} 
\hline
Name & Platform & Type & |Train| & |Test| & Metric\\
\hline \hline
TE Emoji & Twitter & CLS & 45,000 & 50,000 & MA F1\\
TE Emotion & Twitter & CLS & 3,257 & 1,421 & MA F1\\
TE Hate & Twitter & CLS & 9,000 & 2,970 & MA F1\\
TE Irony & Twitter & CLS & 2,862 & 784 & F1\\
TE Offense & Twitter & CLS & 11,916 & 860 & MA F1\\
TE Sentiment & Twitter  & CLS & 45,615 & 12,284 & MA recall\\
TE Stance & Twitter & CLS & 2,620 & 1,249 & non-neutral MA F1 \\
\hline
CCT & CC & CLS & 1,611,934\tablefootnote{Note the discrepancy with TFDS split sizes: we held out 10\% of the TFDS train split for validation and combined the TFDS validation and test splits into a single test split. } & 194,640 &  acc \& F1\\
\hline 
YRP & Yelp & CLS & 522,000 & 38,000 &  acc \& F1\\
\hline 
RTIFU & Reddit & GEN &  71,714 & 3,953  & ROUGE-1\\
\hline
GE & Reddit & CLS & 43,410 & 5,427 &  MA F1 \\
\hline
\end{tabular}
}
\end{table}
\label{sec:benchmark}
Motivated by the distinction between social media language and conventional language observed in \cref{sec:compare}, we compile the Social MedIa Language Evaluation (SMILE\smiley) benchmark that aims to provide a more comprehensive assessment of social media language understanding by improving platform and task coverage. The SMILE\smiley~  benchmark consists of eleven English-only tasks drawing from five datasets that are derived from four source platforms: Twitter, Yelp, Reddit, and Civil Comments\footnote{https://www.drupal.org/project/civilcomments} (a now defunct comment hosting service for web publications). The SMILE\smiley~ benchmark covers both classification and generation tasks in the areas of self-expression, opinion discovery, and online safety. All eleven tasks are publicly available on TensorFlow Datasets\footnote{https://www.tensorflow.org/datasets} (TFDS) as listed in \cref{tab:tfds_names}. In what follows, we describe each dataset (\cref{sec:datasets}) and the summary evaluation metrics (\cref{sec:metrics}) to be used with SMILE. In \cref{sec:similarity}, we investigate whether language in SMILE\smiley~ tasks is more similar to conventional or social media language.
\begin{table}[h]
\centering
 \caption{SMILE benchmark sets in TensorFlow Datasets.}
    \label{tab:tfds_names}
    \resizebox{\columnwidth}{!}{
\begin{tabular}{||l | l l ||} 
\hline
Name & TFDS name & Version \\
\hline \hline
TE Emoji & \texttt{huggingface:tweet\_eval emoji} & \texttt{1.1.0} \\
TE Emotion & \texttt{huggingface:tweet\_eval/emotion} & \texttt{1.1.0} \\
TE Hate & \texttt{huggingface:tweet\_eval/hate} & \texttt{1.1.0} \\
TE Irony & \texttt{huggingface:tweet\_eval/irony} & \texttt{1.1.0} \\
TE Offense & \texttt{huggingface:tweet\_eval/offensive} & \texttt{1.1.0} \\
TE Sentiment & \texttt{huggingface:tweet\_eval/sentiment} & \texttt{1.1.0} \\
TE Stance & \texttt{huggingface:tweet\_eval/stance*} & \texttt{1.1.0}  \\
\hline
CCT & \texttt{civil\_comments} & \texttt{1.2.4} \\
\hline 
YRP & \texttt{yelp\_polarity\_reviews} & \texttt{0.2.0} \\
\hline 
RTIFU & \texttt{reddit\_tifu} & \texttt{1.1.2} \\
\hline
GE & \texttt{goemotions} & \texttt{0.1.0}  \\
\hline
\end{tabular}}
\end{table}
\subsection{Datasets}
\label{sec:datasets}
\cref{tab:benchmarks} summarizes key characteristics of the SMILE\smiley~ datasets. Below we describe each dataset in more detail and discuss the evaluation metrics of each task.

\paragraph{TweetEval (TE, \citet{barbieri2020tweeteval}) } SMILE\smiley~ includes all seven TweetEval classification tasks (emoji prediction, emotion recognition, hate speech detection, irony detection, offensive language identification, sentiment analysis, and stance detection) and uses the same performance metrics for each task as TweetEval. The seven tasks were derived from SemEval\footnote{https://en.wikipedia.org/wiki/SemEval} challenges  between 2016 and 2019.

\paragraph{Civil Comments Toxicity (CCT, \citet{borkan2019nuanced})} This dataset contains news-site user comments collected between 2015 and 2017 labeled for toxicity classification by human raters. A comment is considered toxic if at least one rater labeled it as such. Under this definition, $30\%$ of the comments in the dataset are toxic. SMILE\smiley~ uses accuracy and F1 as the performance measures for this task. 

\paragraph{Yelp Review Polarity (YRP, \citet{zhang2015character}) } Yelp Review Polarity dataset includes Yelp reviews from 2015 labeled positive or negative according to the user star ratings. The dataset is balanced and uses accuracy and F1 as the evaluation metrics. While it is arguable whether online reviews should be considered social media, \cref{sec:similarity} shows that Yelp reviews have more in common with social media language than with conventional language.

\paragraph{Reddit TIFU (RTIFU, \citet{kim2018abstractive}) } RTIFU is a weakly supervised abstractive summarization task derived from the r/tifu subreddit\footnote{https://www.reddit.com/r/tifu/} crawled between 2013 and 2018. The task is to generate a post's title given that post's body content. RTIFU is the only SMILE\smiley~ task suitable for measuring generation capabilities of language models in the social media domain. As the metric for this task, SMILE\smiley~ uses ROUGE-1 score \citep{lin2004rouge}, which ranges between 0 and 1, with higher values indicating better performance.

\paragraph{GoEmotions (GE, \citet{demszky2020goemotions})} The GoEmotions dataset includes snippets from Reddit posts published between 2005 and 2019 labeled with one or more of 27 possible emotional categories. Compared to TweetEval Emotion where each example belongs to only one of four emotional categories, the GoEmotions task involves multiclass classification on a much finer scale. SMILE\smiley~ uses the macro-averaged (MA) F1 score across all categories as the performance metric for this task. 
\subsection{Summary Evaluation Metrics}
\label{sec:metrics}
To measure the overall performance of language models on SMILE we use three summary evaluation metrics that aggregate over the performance score of each the eleven tasks in SMILE\smiley~. The summary metrics are: the task macro-averaged performance, the platform macro-averaged performance, and the TweetEval macro-averaged performance.

\paragraph{Performance Scores} To construct a summary metric, we first compute a performance score for each of the eleven tasks in  \cref{tab:benchmarks}. Note that most tasks only use a single evaluation metric to measure performance. We use these single metrics scaled by $100x$ as their performance scores. For CCT and YRP where both accuracy and F1 are used for evaluation, we compute the average between accuracy and F1 scaled by $100x$ as the performance score following the practice of \citet{wang2018glue, wang2019superglue}. Note that all the evaluation metrics in \cref{tab:benchmarks} range between $0$ and $1$. The overall performance score of each task hence ranges between $0$ and $100$.

\paragraph{Task Macro-Averaged Performance (TMA)} This is the simple average of the performance scores across all the eleven tasks in SMILE. Consequently, this score assumes that each task in SMILE\smiley~ is equally important in measuring the performance of language models for social media language understanding. We refer to this metric as the SMILE\smiley  score. 

\paragraph{Platform Macro-Averaged Performance (PMA)} Because the majority of the SMILE\smiley~ tasks are Twitter-based, it is arguable that the task macro-averaged performance overweights one platform. To mitigate confounding effects of different social media platforms, SMILE\smiley~ also uses a per-platform macro-averaged performance score as a summary evaluation metric. To compute this metric, the performance scores of the tasks from the same platform are first averaged, yielding four per-platform average performance scores. The platform macro-averaged performance is then computed by taking the mean of these four per-platform scores.

\paragraph{TweetEval Macro-Averaged Performance (TEMA)} This summary evaluation metric is a simple average of the performance scores across all the TweetEval tasks. This is the same summary metric used in the TweetEval benchmark. We include this metric as several key existing works \citep{barbieri2022xlm, deluciabernice} evaluate performance on the TweetEval benchmark.

\begin{table}
\centering
 \caption{Difference between the five datasets in SMILE and English mC4 and T\&R}
    \label{tab:similarity}
\begin{tabular}{||c | c c | c c ||} 
\hline
Name &  \multicolumn{2}{c|}{English mC4} & \multicolumn{2}{c||}{T\&R}\\
 & SKL & JD & SKL & JD \\
\hline \hline
All TweetEval & 5.96  & 0.78 & 3.89  & 0.77 \\
\hline
Civil Comments Toxicity & 3.73 & 0.72 & 2.35  & 0.71 \\
\hline 
Yelp Review Polarity & 5.86  & 0.78 & 4.70  & 0.75 \\
\hline 
Reddit TIFU &  5.44  & 0.85 & 4.49  & 0.79 \\
\hline
GoEmotions & 6.03  & 0.83 & 5.91  & 0.76 \\
\hline \hline
\end{tabular}
\end{table}
\subsection{Similarity between SMILE and the Pretraining Corpora}
\label{sec:similarity}
In this section, we seek to understand whether SMILE\smiley~ tasks share more resemblance with social media language than with conventional language. To this end, we compute the difference between five subsets of SMILE\smiley~ and the August 2020 segments of T\&R and the English mC4 corpus introduced in \cref{sec:compare}. Following the protocol established in \cref{sec:compare}, we use SKL divergence and Jaccard distance between token distributions to measure how the language in SMILE\smiley~ tasks differs from T\&R and English mC4. The results are summarized in \cref{tab:similarity}.

From \cref{tab:similarity}, SKL and JD are high between all five SMILE subsets and both T\&R and English mC4. Nonetheless, the five SMILE\smiley~ subsets are less different from T\&R than they are from English mC4 in terms of SKL. This may suggest pretraining on a corpus of social media language as an avenue to improving benchmark performance (as supported by subsequent experiments in \cref{sec:model}). Meanwhile, the high SKL ($>0.25$) could imply that the capacity of the pretrained language models in domain adaptation may still be a key factor in determining downstream benchmark performance. Finally, the comparison results for the Yelp dataset in \cref{tab:similarity} also suggest that Yelp reviews are more similar to social media language than to conventional language, despite user reviews not being considered the prototypical example of social media.

\section{Social Media Domain Adaptation}
\label{sec:model}
In this section, we explore adaptation of language models to the social media domain. We carry out a large scale empirical study of training regimens for producing language models that work well on social media text as proxied by the SMILE benchmark introduced in \cref{sec:benchmark}. Our winning recipe adapts a general T5 language model \citep{raffel2020exploring} by \begin{enumerate*}
\item training a tokenizer on a corpus of both conventional language and social media language and \item pretraining model parameters from scratch, first using the conventional language portion and then the social media language portion.
\end{enumerate*}
The resulting SociAl Media langUage modEL (SAMUEL) can outperform the best alternative of similar size and pretraining budget on the SMILE\smiley~ benchmark by 4.2 points. 

In what follows, we first highlight the key results of comparing SAMUEL with alternative language models in terms of the performance on SMILE\smiley~ (\cref{sec:baselines}). We then provide a detailed description of the empirical study that informed SAMUEL's design. Specifically,
\begin{itemize}[leftmargin=*]
\item In \cref{sec:sm_data}, we introduce a social media language corpus of publicly available social media text that includes data from four major social media platforms (Twitter, Reddit, Facebook, and Telegram).
\item In \cref{sec:tokenization_pretraining}, we show that tokenizer setup and language model pretraining can benefit from the social media language corpus compared to using a corpus of conventional language.
\item We consider the standard domain adaptation practice of continual pretraining in \cref{sec:continual_pretraining}.
\item Finally, we explore strategies for mixing social media and conventional language data as alternatives to continual pretraining (\cref{sec:mixtures}), completing the design of SAMUEL's training recipe.
\end{itemize}
In addition, to understand whether language models adapted for social media understanding can maintain decent performance in general language understanding, we report results on the GLUE \citep{wang2018glue} and SuperGLUE \citep{wang2019superglue} benchmark in the Appendix (\cref{sec:glue-exp}).

\subsection{Performance of SAMUEL and Alternatives on the SMILE Benchmark}
\label{sec:baselines}
We describe SAMUEL and compare its performance on the SMILE benchmark with multiple representative competing language models.
\begin{table}
 \caption{Overall performance on the SMILE\smiley benchmark.}
    \label{tab:baselines_overall}

\begin{tabular}{||c || c c c ||} 
 \hline
 Model & TEMA & PMA & TMA  \\ [0.1ex] 
 \hline\hline
 T5 1.1 base  & 62.26 & 68.22 & 62.39 \\
 BERTweet & 67.90 & n/a & n/a \\
 mT5 base  & 60.86 & 66.89 & 60.62 \\
 byT5 base  & 59.44 & 67.55 & 60.38 \\
  \hline
 SAMUEL & \textbf{67.95} & \textbf{70.75} & 
\textbf{66.63} \\
$\sigma$ & $\pm 0.19$ & $\pm 0.10$ & $\pm 0.14$ \\ 
 
 \hline
 T5 XXL & 68.20 & 71.01 & 66.82 \\
 \hline \hline
\end{tabular}
\end{table}
\paragraph{SAMUEL} SAMUEL is a T5-based language model adapted for social media understanding. SAMUEL is pretrained on a corpus of both social media language and conventional language. We discuss the choice of architecture, training corpus, tokenizer setup, and pretraining procedure for SAMUEL below.
\begin{itemize}[leftmargin=*]
\item \emph{Architecture.}  SAMUEL is a based on the T5 \citep{raffel2020exploring} architecture. It is a transformer-based \citep{vaswani2017attention} encoder-decoder model that is capable of tackling both classification and generation tasks as required by SMILE. It does so by processing the input and output of the tasks as free form text. Following the T5 1.1.\footnote{https://github.com/google-research/text-to-text-transfer-transformer/blob/main/released\_checkpoints.md} implementation, SAMUEL has a parameter size of 220M that matches the size of T5 1.1. base.
\item \emph{Pretraining Corpus.} SAMUEL is pretrained on a corpus that is $80\%$ social media text and $20\%$ conventional web text. The social media language portion is drawn from the corpus described in \cref{sec:sm_data}. The conventional language portion is drawn from the C4 corpus that was used to pretrain the original T5 model.
\item \emph{Tokenizer Setup.} SAMUEL's tokenizer is an SPM with 32k tokens. It is trained directly on the pretraining corpus. As a result, the tokens selected are drawn from both social media and conventional language. 
\item \emph{Pretraining Procedure and Hyperparamters.} SAMUEL is first pretrained on conventional language and then on social media language. For SAMUEL and for other pretraining experiments in this paper, we pretrain all models using the span corruption objective \citep{raffel2020exploring} on tensor processing unit (TPU) pods with batch size of 2048, sequence length of 512,  and a total of $2^{18}$ steps with input packing. We use the Adafactor optimizer with an reciprocal square root decay learning rate schedule.
\end{itemize}

\paragraph{Competing Language Models} We compare five representative language models with SAMUEL on SMILE. Because SAMUEL is T5-based, we consider a variety of other T5-based language models in our comparison. We also consider BERTweet, a state-of-the-art language model that specializes in tweet understanding. We describe each of the competing language models below and provide a summary in \cref{tab:baseline_stats} in the Appendix.
\begin{itemize}[leftmargin=*]
\item \emph{T5 1.1.~base \citep{raffel2020exploring}}  shares the same parameter and tokenizer size as well as the same pretraining hyperparamters as SAMUEL. However, both the model and the tokenizer are trained exclusively on the C4 corpus. It differs from the off-the-shelf version of T5 1.1 base because it was pretrained on 4x as much data to match SAMUEL's pretraining budget.
\item \emph{mT5 base \citep{xue2020mt5}} is an off-the-shelf multilingual variant of T5. It has 580M parameters and is pretrained on mC4 for 1M steps. This model's SPM has 250k tokens including 256 byte tokens, which means no string of text is out of vocabulary for mT5. We explore the byte fallback feature of SentencePiece tokenizers in more detail in \cref{sec:byte_fallback}.
\item \emph{byT5 base \citep{xue2022byt5}} is an off-the-shelf byte-level model that closely follows the T5 setup. Instead of relying on a SentencePiece model for tokenization, byT5 simply uses bytes as tokens to flexibly represent text. We hypothesize that such flexibility may be beneficial for representing text in the social media domain, given its noisy, informal, and dynamic nature. 
\item \emph{T5 1.1 XXL \citep{raffel2020exploring}} is an off-the-shelf large language model with 11B parameters, which is more than 40x larger than SAMUEL. While a head-to-head comparison between T5 XXL and SAMUEL may not be fair because of the discrepancy in parameter size \citep{kaplan2020scaling}, we nonetheless include this model to estimate performance headroom.
\item \emph{BERTweet \citep{nguyen2020bertweet}} is a language model for social media understanding that achieves state-of-the-art performance on the TweetEval benchmark\footnote{https://github.com/cardiffnlp/tweeteval}. BERTweet is a 110M-parameter encoder only model and hence cannot perform generation tasks. We only report its performance on TweetEval.
\end{itemize}

\paragraph{Protocol} After pretraining SAMUEL, we fine-tune SAMUEL and the five alternative models on each SMILE\smiley~ task separately for 10k steps and report the performance metrics on the test splits. To get a sense of the uncertainty in performance metrics, we fine-tune SAMUEL on each task 12 times and compute the standard deviation for each metric. Note that we presume the performance of different models to share similar levels of uncertainty and do not compute the standard deviation for each model because of the high computational cost of doing so. We run all experiments using the \verb|T5X| framework \cite{t5x2022}.

\paragraph{Results} \cref{tab:baselines_overall} summarizes the performance comparison of SAMUEL against other language models on SMILE. With the exception of T5 XXL, SAMUEL outperforms its  similar-sized alternatives by at least 4.2 points in terms of task macro-averaged (TMA) score. This suggests the practical utility of the recipe behind SAMUEL in building effective language models for social media language understanding. Moreover, SAMUEL's score is only 0.2 points below that of T5 1.1 XXL with 50 times more parameters, which suggests that SAMUEL may be nearing the upper limits of performance headroom on this benchmark.

\begin{table}
 \caption{Summary performance on SMILE when SPM tokenizer learning and/or LM pretraining are/is conducted on the SM corpus vs the C4 corpus.}
    \label{tab:ablations_overall}
\begin{tabular}{||l || c c c ||}  
 \hline
 Model & TEMA & PMA & TMA  \\ [0.1ex] 
 \hline\hline
 T5 1.1 base & 62.26 & 68.22 & 62.39 \\
 \hline \hline
 + SM SPM  & +2.61 & +1.05 & +1.83 \\
+ SM Pretraining & +3.68 & +1.70 & +2.77 \\
+ SM SPM\&Pretraining  & \textbf{+5.40} & \textbf{+2.00} & \textbf{+3.73} \\

 \hline \hline
Continual Pretraining & +2.81 & +1.58 & +2.27 \\
 \hline 
\hline
\end{tabular}
\end{table}
\begin{table*}
\centering
\caption{Performance on each task in SMILE when SPM tokenizer learning and/or LM pretraining are/is conducted on the SM corpus vs the C4 corpus.}
\label{tab:ablations_per_task}
\begin{tabular}{|| l || c c c c c c c | c c | c c | c | c  ||} 
 \hline
 Model & \multicolumn{7}{c|}{TweetEval} & \multicolumn{2}{c|}{CCT}	& \multicolumn{2}{c|}{YRP}	& RTIFU &	GE  \\ [0.1ex] 
 & Emoj & Emot & H & I & O & Sen & Sta & Acc & F1 & Acc & F1 & R-1 & F1  \\ 
 \hline\hline
 T5 1.1 base & 31.25 & 75.65 & 40.99 & 69.96 & 77.68 & 72.33 & 67.98 & 81.58 & 65.19 & 97.39 & 97.39 & 28.79 & 50.91 \\
 \hline \hline
 + SM SPM & +3.98 & +6.13 & +0.82 & +2.16 & +1.90 & +0.85 & +2.44 & -0.12 & +2.50 & +0.10 & +0.11 & -0.29 & +0.86 \\
+ SM Pretraining & +4.91 & +3.89 & +6.18 & +5.07 & +1.07 & +1.67 & +2.99 & -0.53 & \textbf{+3.53} & +0.12 & +0.12 & +0.25 & +2.78  \\
 + SM SPM\&Pretraining & \textbf{+5.73} & \textbf{+8.15} & \textbf{+8.88} & \textbf{+8.34} & +0.86 & \textbf{+1.88} & \textbf{+3.93} & \textbf{+0.15} & +3.31 & \textbf{+0.29} & \textbf{+0.30} & -0.25 & +1.43 \\
 
 \hline \hline
Continual Pretraining  & +2.81 & +4.13 & +5.20 & +1.71 & \textbf{+2.18} & +1.23 & +2.40 & -0.26 & +3.43 & +0.14 & +0.16 & \textbf{+0.26} & \textbf{+3.32} \\
 \hline \hline
\end{tabular}
\end{table*}
\subsection{Social Media Corpus}
\label{sec:sm_data}
In \cref{sec:compare}, we observe that  social media language is different from conventional language. Meanwhile, commonly used pretraining corpora, like C4 in the case of off-the-shelf T5, represent conventional language. We begin our adaptation of T5 to social media language understanding with compiling a large social media (SM) corpus. 

\paragraph{Protocol} We parse out post text from publicly crawlable Twitter, Reddit, Facebook, and Telegram pages. We filter out posts with images, videos, or URLs because the text of such posts may not contain the entirety of their meaning and therefore may be too difficult to learn from in the span corruption setting. To mitigate the risk of inadvertently training any models on posts from the benchmark test sets, we only include posts published in 2022 to avoid temporal overlap (the content used in SMILE\smiley~ was published between 2015 and 2020). 

\paragraph{Results} The resulting corpus has 6B examples. Its platform composition is 51\% Twitter, 26\% Reddit, 23\% Facebook, and 0.1\% Telegram. In contrast, the C4 corpus contains only $0.005\%$ Reddit data and no data from the other three platforms.  In addition, while C4 has 16 times fewer examples, C4 examples are on average 10 times longer than the examples in our social media corpus. The presence of short examples makes it important to minimize padding with input packing. 

\subsection{In-Domain Tokenization and Pretraining}
\label{sec:tokenization_pretraining}
We consider the effect of using the SM corpus defined in \cref{sec:sm_data} for tokenizer setup and LM pretraining on the model's capacity for social media language understanding.

\paragraph{Protocol} Keeping the T5 1.1 backbone fixed, we modify its tokenizer and pretraining regimen separately and then together.  We measure the impact on the SMILE\smiley~ score.

\paragraph{Results} \cref{tab:ablations_overall} and \cref{tab:ablations_per_task} show the metric improvement compared to the baseline model---a T5 1.1 model where both the tokenizer and the model are trained on the C4 corpus.  Unsurprisingly, including in-domain data helps the performance on most tasks. Swapping out the standard tokenizer with one trained on SM data and then pretraining on C4 leads to a 1.8-point gain on the overall SMILE\smiley~ score (see \cref{tab:example} in the Appendix for examples), while using SM pretraining data with a standard tokenizer adds 2.8 points. These performance gains are not exactly additive when both the model's tokenizer and parameters are trained on the in-domain data, but the combination does lead to the biggest improvement---3.7 points to the TMA. It is worth noting the tasks that buck the overall trend. Reddit TIFU, the only summarization task, appears to be hurt by a tokenizer trained on in-domain data, whether or not the rest of the model is pretrained on SM or C4. In-domain pretraining with a C4 tokenizer does improve the ROUGE-1 score, but barely more than a single standard deviation for this task (+0.25 vs 0.22), all despite the fact that the token distribution of this dataset is closer to SM data than it is to clean web text (\cref{tab:similarity}). The opposite effect can be observed for the TweetEval Emotion task---there, the in-domain tokenizer alone yields a 6.1 point gain, 1.5 times larger than from in-domain pretraining.

\begin{figure*}
    \centering
    \includegraphics[width=1.0\linewidth]{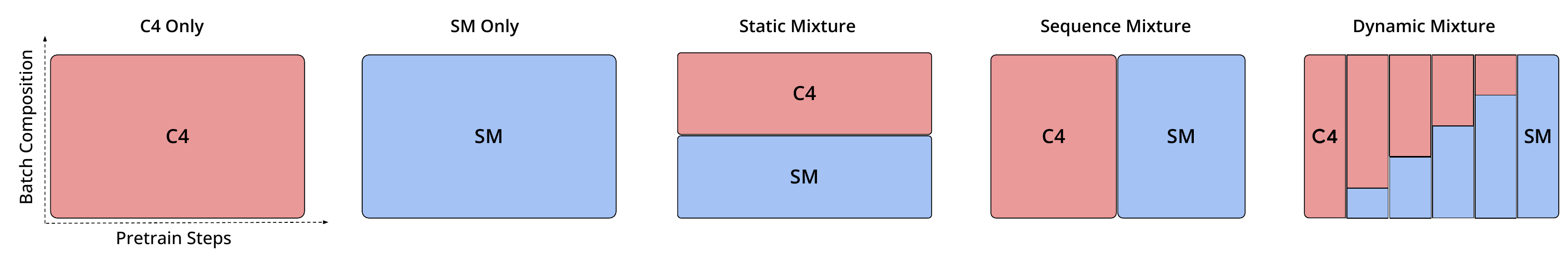}
    \vspace{-1em}
    \caption{Representation of pretrain dataset composition in terms of the mix of SM and C4 dataset as a function of pretraining steps. Two leftmost diagrams: SM and C4 only pretrain datasets. Three rightmost diagrams: the three types of mixing schedules: static, sequence, and dynamic.}
    \label{fig:mixture}
    \vspace{-1.5em}
\end{figure*}
\subsection{In-Domain Continual Pretraining}
\label{sec:continual_pretraining}
While in \cref{sec:tokenization_pretraining} we established that using social media data in tokenizer and model training significantly improves performance on the SMILE\smiley~ benchmark, this recipe requires retraining the model from scratch, which is undesirable in practice because of engineering effort, data, and computation required. A much more popular and practical recipe is continual pretraining, where an existing pretrained model is adapted to a different domain or task via more pretraining steps. In this section, we investigate how well continual pretraining can perform in the social media domain.

\paragraph{Protocol} We take a T5 model with a standard C4 tokenizer pretrained on C4 for $2^{17}$ steps and continue training it on the same span corruption task but on our four-platform SM corpus for another $2^{17}$ steps, yielding a total of $2^{18}$ pretraining steps. 

\paragraph{Results}   \cref{tab:ablations_overall} and \cref{tab:ablations_per_task} report  the results of our continual pretraining experiment. We see that continual pretraining does not help as much as training a T5 1.1 backbone on SM from scratch (+2.3 vs +2.8). This finding is in line with \citet{gu2021domain}. However, it is remarkable that continual pretraining achieves $80\%$ of the gain afforded by from-scratch in-domain pretraining---while using half as much in-domain data, in half as many steps. Part of the reason may be that the model is able to leverage the foundation derived from the clean language of its original pretraining corpus, C4, despite the fact that it is much further from social media language in token distribution.

\subsection{Mixed-Domain Pretraining}
\label{sec:mixtures}
Inspired by the relative success of the continual pretraining recipe in \cref{sec:continual_pretraining}, we drill down into the idea of combining in- and out-of-domain data into mixed-domain data to further improve the performance. In particular, we consider various mixing schedules and ratios.

\subsubsection{Mixing Schedules}
\paragraph{Protocol} We try three different mixing schedules for C4 and our SM corpus (see \cref{fig:mixture}):
\begin{enumerate*}
    \item Static mixture: every batch contains the same proportion of C4 and SM data.
    \item Sequential mixture: in the first part of training, batches consist of only C4, in the second part---only SM. This is similar to continual pretraining.
    \item Dynamic mixture: in the beginning of training batches contain mostly C4 examples. As the training progresses, the proportion of C4 decreases and the proportion of SM increases linearly. By the end of the training, batches contain mostly SM examples. 
\end{enumerate*}
As we learned from \cref{sec:tokenization_pretraining}, updating the tokenizer to match the target distribution contributes significantly to the model performance. We train a tokenizer on a $50/50$ mix of C4 and SM to match the pretraining data and pretrain three models using the schedules described above.

\paragraph{Result} \cref{tab:mixtures} shows that the model trained on the sequential mixture performs best. Note that though its schedule is identical to continual pretraining, the model is different because it has a custom tokenizer and was pretrained from scratch. We also note that the dynamic mixture is very close in performance to sequential: 66.55 vs 66.38 on the overall score, which has a standard deviation of 0.14. However, pretraining with a dynamic schedule is much more challenging to set up from the infrastructure perspective than the sequence mixture. We hence commit to the use of sequential mixture schedule in our subsequent investigation.

\subsubsection{Mixing Rates}
\paragraph{Protocol} Using the sequence mixture as the schedule, we then try three different SM/C4 ratios: 20/80, 50/50 and 80/20. We update the tokenizers to match the ratios and pretrain two more models. 
\paragraph{Results} The best overall model is trained on 20\% C4 and 80\% SM in sequence, with an overall score of 66.63. (That corresponds to our best model, SAMUEL, in \cref{tab:baselines_overall}). Interestingly, different benchmark tasks react to the rate of in-domain data differently (see \cref{tab:mixtures_per_task} in the Appendix). For TweetEval Emotion, the more in-domain data, the better---its performance peaks when the model is trained exclusively on SM data. On the other hand, Yelp polarity classification benefits roughly equally from any proportion of in-domain pretraining, while Reddit TIFU summarization benefits from more C4 pretraining, even though it is out of domain.

\begin{table}[t]
 \caption{Overall performance on SMILE using different strategies and ratios to mix the SM corpus and the C4 corpus. *indicates the configuration adopted by SAMUEL.}
    \label{tab:mixtures}
\begin{tabular}{|| c || c || c c c ||}  
 \hline
 Strategy & C4/SM & TEMA & PMA & TMA \\ [0.1ex] 
\hline\hline
Static & 50/50 & 67.31 &	70.18 &	65.96 \\
Sequential & 50/50 & \textbf{68.28}	& 70.42	& 66.55 \\
Dynamic & 50/50 & 67.87	& 70.44	& 66.38 \\
\hline
None & 0/100 & 62.26 &	68.22 & 62.39 \\
Sequential & 20/80 & 67.83	& 70.35	& 66.30 \\
Sequential & 50/50 & \textbf{68.28}	& 70.42	& 66.55 \\
*Sequential & 80/20 & 67.95 &	\textbf{70.75}	& \textbf{66.63} \\
None & 100/0 & 67.66 & 70.22 & 66.12 \\
  \hline
 \hline
\end{tabular}
\end{table}
\section{Ablation Studies}
\label{sec:ablation}
We conduct ablation studies on the platform composition of our SM corpus and the effect of the byte-level fallback feature of SPM tokenizers to see how these aspects impact social media language understanding.
\begin{itemize}[leftmargin=*]
\item In \cref{sec:xplatform}, we evaluate performance contributions of using data from different SM platform and conclude that a variety of platforms in the training data yields optimal performance on the SMILE benchmark.
\item In \cref{sec:byte_fallback}, we consider adding byte tokens to the model vocabulary, which should intuitively help parse noisy social media text, and find that while byte fallback does indeed help the model trained on conventional language, it no longer has a positive effect when the model is trained on SM data.
\end{itemize}

\subsection{Cross-Platform Transfer}
\label{sec:xplatform}
\begin{table}[t]
 \caption{Performance of LMs pretrained only on data from a certain platform vs all platforms.}
    \label{tab:platform_transfer}
\centering
\resizebox{\columnwidth}{!}{
\begin{tabular}{|| c || c c c c | c||}  
 \hline
 PT data & Twitter avg & Reddit avg & CC avg	& Yelp avg & PMA \\ [0.1ex] 
\hline\hline
All SM & 67.66 & \textbf{40.44} & \textbf{75.12} & 97.69 & \textbf{67.60} \\
  \hline
Twitter & \textbf{68.12} & 39.89 & 75.02 & 97.42 & 67.54 \\
Reddit & 66.80 & 40.05 & 75.04 & \textbf{97.87} & 67.38 \\
Facebook & 66.23 & 40.35 & 74.99 & 97.40 & 67.11 \\
  \hline
 \hline
\end{tabular}
}
\end{table}

Given the multi-platform nature of the SMILE\smiley~ benchmark, we investigate how using the four-platform mixture compares to training the model on each platform separately. For example, we presume Reddit data will be helpful on the Reddit-based tasks, but how well will it transfer to tasks from other platforms?
\paragraph{Protocol} To answer this question, we train three more T5 models using the best performing recipe from \cref{sec:tokenization_pretraining} but on single-platform slices of the same SM corpus --- Twitter-only, Reddit-only, and Facebook-only. The amount of Telegram data in the original SM corpus is negligibly low---about $0.1\%$ and only 7M examples---so we can not train a Telegram-only model. The three single-platform models see the same amount of data despite different average example lengths thanks to input packing.

\paragraph{Results} The results for per-platform average performance scores as well as the PMA are shown in \cref{tab:platform_transfer}. Overall, we see that single-platform models perform very similarly to each other as well as the four-platform mixture ("All SM"). Though the four-platform mixture has the highest score on the PMA, it is within one standard deviation from the second place---the Twitter-only model. The Twitter-only model also, unsurprisingly, does best on the Twitter portion of the SMILE\smiley~ benchmark. On the other hand, the Reddit-only model takes the third place on the Reddit portion of the benchmark, lagging behind the four-platform mixture as well as the Facebook-only model. We conclude that though the scores are very close, having a mix of platforms in the pretraining data results in beneficial transfer across platforms.

\subsection{Byte-level Fallback}
\label{sec:byte_fallback}
\begin{table}
 \caption{Performance of LMs when pretrained with/without byte-level fallback and with/without in-domain data and SPM tokenizer.}
    \label{tab:byte_fb}
\resizebox{\columnwidth}{!}{
\begin{tabular}{||l || c c c ||} 
 \hline
 Model & TEMA & PMA & TMA  \\ [0.1ex] 
 \hline\hline
 T5 1.1 & 62.26 & 68.22 & 62.39 \\
\hline
 + Byte Fallback  & +1.20 & +0.56 & +0.87 \\
+ SM SPM\&Pretraining  & \textbf{+5.40} & +2.00 & \textbf{+3.73} \\
+ SM SPM\&Pretraining + Byte Fallback  & +4.83 & \textbf{+2.01} & +3.47 \\
\hline \hline
\end{tabular}
}
\end{table}
We evaluate the impact of including byte tokens in the model vocabulary. Also known as byte-level fallback, this feature allows the tokenizer to segment any string of text without resorting to the \verb|<UNK>| token. Can it improve performance on social media text with its typos, abbreviations, and emojis?

\paragraph{Protocol} To answer this question, we replace the 256 least frequent tokens in the standard C4 SPM with tokens representing the 256 bytes and use the modified tokenizer to pretrain a model on C4. We perform the same vocabulary surgery on a model whose tokenizer and parameters are trained on social media data.

\paragraph{Results} The results are shown in \cref{tab:byte_fb}. Overall, even though the C4 model with byte fallback did not see a single example from the social media corpus, its performance on the SMILE\smiley~ benchmark is 0.87 points ($\sigma = 0.14$) better than the equivalent model without byte fallback. We speculate that allowing the tokenizer to fall back to byte tokens does indeed help the model handle emojis and typos that  otherwise would not be represented by the vocabulary. However, for the model trained on social media data, replacing 256 organically selected tokens with bytes no longer has a positive effect (+3.73 vs +3.47).

\section{Future Work}
\label{sec:future_work}
We envision the following directions as future work.
\begin{itemize}[leftmargin=*]
\item \emph{Internationalization.} Our SMILE\smiley~ benchmark spans multiple platforms and tasks but is currently English only. An important opportunity for future work is the extension to other languages, ideally covering locales where social media usage is most pronounced and locales with multilingual content.
\item \emph{Multi-modality.} Social media content is inherently multi-modal and context dependent. Relevant modalities can be content based (text, image, etc.) but also include creators, communities, and any other aspects of a social graph. Future opportunities lie both in exploring techniques for modeling multi-modal content and in providing evaluation benchmarks for this setting.
\item \emph{Time Sensitivity and Adaptability.} Social media language changes faster than conventional language (\cref{sec:v-shift}). Consequently, it would be valuable to define time-sensitive tasks that measure the ability of models to adapt to these shifts. Byte-level language models \citep{tay2021charformer, xue2022byt5} may be of particular interest in tackling distribution shifts because they are not constrained by a fixed vocabulary and can be fully fine-tuned (i.e. including the tokenizer). Efficient architectures \citep{tay2022efficient} that hinge on the principle of sparsity \citep{dao2022flashattention,  geng2020joint, geng2019partially, geng2018stochastic, geng2018temporal, geng2017efficient, kuang2017screening, paria2020minimizing}, among other principles \citep{jaegle2021perceiver, choromanski2020rethinking, poli2023hyena}, can be particularly relevant given the longer sequence length induced by byte representation of the data.

\item \emph{Domain Adaptation vs. Model Scaling.} Given the popularity of very large language models, it would be beneficial to study how scaling up the model size affects the gains from domain adaptation. At what model size, if any, does the performance improvement afforded by domain adaptation cease to be worth the computationally expensive pretraining regimen?
\end{itemize}

\section{Conclusion}
\label{sec:conclusion}
We have shown that the language of social media is significantly different from the standard language of the web and proposed the SMILE\smiley~benchmark that captures the peculiarities of this distinct domain.
Through pretraining experiments with T5, we came up with simple recipes for improving performance on the benchmark.  
We believe it can open up opportunities for future research in adapting language models to the social media domain.

\balance
\bibliography{references}

\onecolumn

\section{Appendix}

In the appendix, we include additional experiment details (\cref{sec:exp-details-appendix}) as well as evaluation results on GLUE and SuperGLUE (\cref{sec:glue-exp}). We also provide concrete examples demonstrating the potential benefits of training the tokenizer and the model parameters on social media data compared to their generic counterparts in \cref{sec:tokenization-example-appendix}.

\subsection{Additional Experiment Details}
\label{sec:exp-details-appendix}

\begin{table}[H]
\centering
 \caption{Key characteristics of SAMUEL and its competing LMs}
    \label{tab:baseline_stats}
\begin{tabular}{||c || c c c c c | c c c ||} 
 \hline
 Model & \multicolumn{5}{c|}{Pretraining} & \multicolumn{3}{c||}{Tokenizer}  \\ [0.1ex] 
 & corpus & \# params & steps & batch & seq length & corpus & \# tokens & has bytes \\ 
 \hline\hline
 T5 1.1 base & C4 & 220M & 262k & 2048 & 512 & C4 & 32k & no \\
 BERTweet & tweets & 110M & 950k & 7k & 128 & tweets & 64k & no \\
 mT5 base & mC4 & 580M & 1M & 1024 & 1024 & mC4 & 250k & yes \\
 byT5 base & mC4 & 580M & 1M & 1024 & 1024 & mC4 & 256 & yes \\
  \hline
 SAMUEL & 20/80 C4/SM & 220M & 262k & 2048 & 512 & 20/80 C4/SM & 32k & no \\
 \hline
 T5 XXL & C4 & 11B & 1M & 2048 & 512 & C4 & 32k & no \\
 \hline \hline
\end{tabular}
\end{table}
\begin{table}[H]
\centering
 \caption{Performance of SAMUEL and its competing models on each task in SMILE.}
    \label{tab:baselines_per_task}
\begin{tabular}{||c || c c c c c c c | c c | c c | c | c ||} 
 \hline
 Model & \multicolumn{7}{c|}{TweetEval} & \multicolumn{2}{c|}{CCT}	& \multicolumn{2}{c|}{YRP}	& RTIFU &	GE  \\ [0.1ex] 
 & Emoj & Emot & H & I & O & Sen & Sta & Acc & F1 & Acc & F1 & R-1 & F1  \\ 
 \hline\hline
 T5 1.1 base & 31.25 & 75.65 & 40.99 & 69.96 & 77.68 & 72.33 & 67.98 & 81.58 & 65.19 & 97.39 & 97.39 & 28.79 & 50.91 \\
 BERTweet & 33.4 & 79.3 & \textbf{56.4} & \textbf{82.1} & \textbf{79.5} & 73.4 & 71.2 & n/a & n/a & n/a & n/a & n/a & n/a \\
 mT5 base & 30.63 & 72.33 & 46.32 & 63.94 & 77.23 & 69.17 & 66.38 & 82.1 & 68.24 & 97.42 & 97.4 & 23.24 & 45  \\
 byT5 base & 32.42 & 67.24 & 44.37 & 60.55 & 74.95 & 67.24 & 69.32 & \textbf{83.04} & 68.67 & \textbf{97.60} & \textbf{97.58} & 27.32 & 47.33  \\
  \hline
 SAMUEL & \textbf{37.42} & \textbf{83.52} & 50 & 78.7 & 78.49 & \textbf{74.35} & \textbf{73.15} & 81.74 & \textbf{68.69} & 97.56 & 97.55 & \textbf{29} & \textbf{55.55}   \\
  $\sigma$ & $\pm 0.25$ & $\pm 0.24$ & $\pm 0.72$ & $\pm 0.56$ & $\pm 0.31$ & $\pm 0.13$ & $\pm 0.41$ & $\pm 0.37$ & $\pm 0.51$ & $\pm 0.01$ & $\pm 0.01$ & $\pm 0.22$ & $\pm 0.45$ \\ 
 \hline
 T5 XXL & 35.42 & 82.45 & 51 & 80.96 & 78.97 & 71.61 & 77 & 82.65 & 68.48 & 98.57 & 98.57 & 32.93 & 50.51 \\
 \hline \hline
\end{tabular}
\end{table}
\begin{table}[H]
\centering
 \caption{Effect of pretraining data mix ratio on each task in SMILE}
    \label{tab:mixtures_per_task}
\begin{tabular}{||l c || c c c c c c c | c c | c c | c | c ||} 
 \hline
 Strategy & C4/SM & \multicolumn{7}{c|}{TweetEval} & \multicolumn{2}{c|}{CCT}	& \multicolumn{2}{c|}{YRP}	& RTIFU &	GE  \\ [0.1ex] 
 & & Emoj & Emot & H & I & O & Sen & Sta & Acc & F1 & Acc & F1 & R-1 & F1  \\ 
\hline\hline
T.5 1.1 base & 0/100 & 31.25 & 75.65 & 40.99 & 69.96 & 77.68 & 72.33 & 67.98 & 81.58 & 65.19 & 97.39 & 97.39 & 28.79 & 50.91 \\
\hline
Sequential & 20/80  & +4.76 & +7.36 & +8.60 & +8.54 & +1.28 & \textbf{+2.20} & \textbf{+6.24} & +0.09 & +3.28 & +0.20 & +0.19 & \textbf{+0.69} & +1.40 \\
Sequential & 50/50  & +5.59 & +7.68 & \textbf{+9.21} & \textbf{+9.99} & \textbf{+2.44} & +1.93 & +5.31 & -0.27 & \textbf{+3.62} & +0.23 & +0.24 & +0.23 & +1.49 \\
*Sequential & 80/20 & \textbf{+6.17} & +7.87 & +9.01 & +8.74 & +0.81 & +2.02 & +5.17 & \textbf{+0.16} & +3.50 & +0.17 & +0.16 & +0.21 & \textbf{+4.64} \\
None & 100/0        & +5.73 & \textbf{+8.15} & \textbf{+8.88} & +8.34 & +0.86 & +1.88 & +3.93 & +0.15 & +3.31 & \textbf{+0.29} & \textbf{+0.30} & -0.25 & +1.43 \\
  \hline
 \hline
\end{tabular}
\end{table}
\subsection{GLUE and SuperGLUE Experiments}
\label{sec:glue-exp}
To understand performance on standard NLU tasks, we report detailed results on the GLUE (\citep{wang2018glue} and SuperGLUE (\citep{wang2019superglue}) benchmarks in \cref{tab:glue-results} and \cref{tab:superglue-results} respectively. All models were pretrained for $2^{18}$ steps, including our version of T5 1.1 (which thus is different from the publicly available version). We trained each model on a mixture of all tasks with proportionate sampling (GLUE and SuperGLUE separately) for 50k steps with a batch size of 128 and a dropout rate of 0.1.
\begin{table}[h]
    \large
    \centering
    \caption{GLUE Results}
    \label{tab:glue-results}
\resizebox{\textwidth}{!}{
    \begin{tabular}{|| l c c | c c c c c c c c c c c c | c ||}
    \hline
     & Pretraining & Tokenizer & COLA & SST2 & MRPC & MRPC & STS-B & STS-B & QQP & QQP & MNLI-m & MNLI-mm & QNLI  & RTE & \\
     [0.1ex]
     & & & Matthew's & Acc & Acc & F1 & Pearson & Spearman & Acc & F1 & Acc & Acc & Acc & Acc & Macro Avg \\
    \hline
    T5 1.1 Base (our version) & C4 & C4 & 50.46 & \textbf{94.38} & 87.25 & 90.61 & \textbf{89.74} & \textbf{89.64} & 91.52 & 88.54 & 87.08 & 86.50 & \textbf{91.82} & 76.53 & 83.82 \\
    & SM & SM & 49.86 & 93.58 & 88.97 & 92.06 & 89.09 & 89.07 & 91.61 & 88.67 & 86.52 & 86.43 & 90.92 & 71.12 & 82.79 \\
    SAMUEL & C4/SM 20/80 & C4/SM 20/80 & 45.17 & 94.15 & 88.48 & 91.77 & 88.90 & 88.72 & 91.63 & 88.73 & 87.09 & 86.96 & 91.85 & 74.37 & 82.94 \\
    & Twitter & Twitter & 46.65 & 93.46 & \textbf{89.95} & \textbf{92.72} & 88.88 & 88.75 & 91.37 & 88.39 & 85.35 & 85.53 & 90.30 & 71.48 & 82.08  \\
    & Reddit & Reddit & \textbf{52.10} & \textbf{94.38} & 89.46 & 92.34 & 89.26 & 89.15 & \textbf{91.70} & \textbf{88.95} & \textbf{87.10} & \textbf{87.52} & 91.29 & \textbf{76.90} & \textbf{84.13} \\
    & Facebook & Facebook & 32.21 & 92.09 & 88.48 & 91.77 & 88.93 & 88.90 & 91.36 & 88.12 & 85.27 & 85.51 & 90.76 & 69.68 & 80.12\\
    \hline 
    \end{tabular}}
\end{table}

\begin{table}[h]
    \large
    \centering
    \caption{SuperGLUE Results}
    \label{tab:superglue-results}
\resizebox{\textwidth}{!}{
    \begin{tabular}{|| l c c | c c c c c c c c c c | c ||}
    \hline 
     & Pretraining & Tokenizer & BoolQ & CB & CB & COPA & MultiRC & MultiRC & ReCoRD & RTE & WiC & WSC &  \\
     [0.1ex]
     & & & Acc & Acc & F1 & Acc & F1 & EM & EM & Acc & Acc & Acc & Macro Avg \\
    \hline
    T5 1.1 Base (our version) & C4 & C4 & 65.87 & 71.43 & 49.84 & 45.00 & \textbf{71.30} & 19.83 & \textbf{74.56} & 68.59 & 63.48 & \textbf{76.92} & 62.58 \\
    & SM & SM & 66.79 & 58.93 & 36.71 & \textbf{55.00} & 66.16 & 18.47 & 66.51 & 65.70 & \textbf{64.73} & 70.19 & 59.88 \\
    SAMUEL & C4/SM 20/80 & C4/SM 20/80 & 66.39 & \textbf{80.36} & \textbf{56.15} & 49.00 & 69.28 & 20.25 & 69.23 & 67.15 & 63.64 & 73.08 & \textbf{62.69} \\
    & Twitter & Twitter & 63.76 & 60.71 & 39.59 & 45.00 & 60.99 & 16.16 & 63.61 & 67.87 & 62.85 & 69.23 & 57.63 \\
    & Reddit & Reddit & \textbf{68.65} & 78.57 & 54.80 & 50.00 & 70.62 & \textbf{20.46} & 68.99 & \textbf{69.68} & 63.17 & \textbf{76.92} & 62.42 \\
    & Facebook & Facebook & 63.24 & 69.64 & 48.68 & 54.00 & 67.01 & 17.31 & 66.32 & 61.01 & 63.95 & 69.23 & 59.88 \\
    \hline
    \end{tabular}}
\end{table}
\subsection{Examples of Social Media Language Tokenization}
\label{sec:tokenization-example-appendix}
 In \cref{tab:example} we present three real-world examples below to illustrate the misalignment between conventional and social media language as well as the resulting difference in predictions made by the T5 1.1. baseline and SAMUEL (trained on conventional and social media language, respectively). These examples are drawn from the TweetEval 4-way Emotion classification task.

\begin{table}[h] 
\fontsize{8pt}{9pt}\selectfont
    \centering 
    \captionsetup{justification=centering}
    \caption{Example tweets from the TweetEval Emotion task processed by T5 1.1 and SAMUEL.}
    \label{tab:example}
\begin{tabular}{||p{0.2cm} | >{\raggedright\arraybackslash}p{2.7cm} || >{\raggedright\arraybackslash}p{6.2cm} | >{\raggedright\arraybackslash}p{6.2cm} ||}  
 \hline
 \# & Tweet with emotion label & T5 1.1 tokenization and prediction & SAMUEL tokenization and prediction
\\ [0.1ex] 
 \hline \hline
 1 & \texttt{"When you baby has their first temperature and all you do is worry \#firsttimemum \#firsttimemom \#newborn \#baby \#sickbaby \#worry"} &
\texttt{[▁When] [▁you] [▁baby] [▁has] [▁their] [▁first] [▁temperature] [▁and] [▁all] [▁you] [▁do] [▁is] [▁worry] [▁\#] [first] [time] [m] [um] [▁\#] [first] [time] [m] [o] [m] [▁\#] [new] [born] [▁\#] [bab] [y] [▁\#] [s] [ick] [bab] [y] [▁\#] [w] [or] [ry]} 

&
\texttt{[▁] [When] [▁you] [▁] [baby] [▁] [has] [▁] [their] [▁first] [▁] [temperature] [▁] [and] [▁] [all] [▁you] [▁] [do] [▁] [is] [▁] [worry] [▁\#] [first] [time] [mum] [▁\#] [first] [time] [mom] [▁\#] [newborn] [▁\#] [baby] [▁\#] [sick] [baby] [▁\#] [worry]} 

\\
& \textbf{Label: sadness} & \textbf{Prediction: optimism} & \textbf{Prediction: sadness}
\\
\hline
 2 & \texttt{"I thought he cried over some of his relative death or something but when i know the truth . I just wanna burst out \laughcry{}"}
 & \texttt{[▁I] [▁thought] [▁] [he] [▁] [cried] [▁over] [▁some] [▁of] [▁his] [▁relative] [▁death] [▁or] [▁something] [▁but] [▁when] [▁] [i] [▁know] [▁the] [▁truth] [▁] [.] [▁I] [▁just] [▁wann] [a] [▁bur] [s] [t] [▁out] [▁] [<UNK>]}
 & \texttt{[▁] [I] [▁thought] [▁] [he] [▁] [cri] [ed] [▁over] [▁] [some] [▁] [of] [▁] [his] [▁] [relative] [▁death] [▁] [or] [▁] [something] [▁] [but] [▁] [when] [▁] [i] [▁know] [▁] [the] [▁] [truth] [▁] [.] [▁] [I] [▁] [just] [▁] [wanna] [▁] [burst] [▁] [out] [▁] [\laughcry{}]}

 \\
 &\textbf{Label: joy} & \textbf{Prediction: sadness} & \textbf{Prediction: joy} \\
 \hline
3 & \texttt{"Actually gunna miss America a lot \anxious{}"} & 
\texttt{[▁Actually] [▁gun] [n] [a] [▁miss] [▁America] [▁] [a] [▁lot] [▁] [<UNK>]} & 
\texttt{[▁] [Actual] [ly] [▁] [gun] [na] [▁] [miss] [▁America] [▁] [a] [▁] [lot] [▁] [\anxious{}]} 

\\
& \textbf{Label: sadness} & \textbf{Prediction: joy} & \textbf{Prediction: sadness} \\
\hline \hline
\end{tabular}
\end{table}
\paragraph{On interpreting tokenization results} Each group of symbols in square brackets represents a single token. There are two special symbols: "\texttt{▁}", which represents a word-separating character like a space, and "\texttt{<UNK>}", which represents an out-of-vocabulary token the tokenizer resorts to when it cannot break down the input string into any of its known 32k tokens.
\paragraph{Discussion} As one can see in \cref{tab:example}, T5 1.1. struggles with hashtag tokenization and fails to understand semantically important emojis. In example 1, T5's tokenizer (trained on C4's clean text) tokenizes properly space-separated text well (e.g. \texttt{[▁worry]}) but when the same text appears inside a hashtag, it gets mangled (\texttt{[▁\#] [w] [or] [ry]}). In contrast, SAMUEL’s tokenizer is capable of tokenizing hashtags in a manner that better aligns with human intuition because it learns not to assume that words will be properly space-separated (\texttt{[▁\#] [worry]}). SAMUEL’s tokenizer is also capable of identifying semantically important emojis (\laughcry{} and \anxious{} in examples 2 and 3), while T5 does not because they are unknown to its tokenizer. These examples show that the token distribution differences reflect real distinctions in language use between social media and the standard web. In addition, the vocabulary learned from these distinct distributions influences downstream task performance: in all three examples, SAMUEL predicts the correct labels while T5 1.1. fails to do so.
\end{document}